\pdfoutput=1

\documentclass[11pt]{article}

\usepackage[final]{acl}

\usepackage{times}
\usepackage{latexsym}

\usepackage[T1]{fontenc}

\usepackage[utf8]{inputenc}

\usepackage{microtype}

\usepackage{inconsolata}

\usepackage{graphicx}

\usepackage{microtype}
\usepackage{hyperref}
\usepackage{url}
\usepackage{booktabs}

\usepackage{amsmath}
\usepackage{mdframed} %
\usepackage{graphicx} %
\usepackage{float}
\usepackage{booktabs} %
\usepackage{multirow} %
\usepackage{comment} %
\usepackage{amssymb} %
\usepackage{multicol}
\usepackage{dsfont} %
\usepackage{makecell} %
\usepackage{enumerate} %
\usepackage[export]{adjustbox}
\usepackage{tabularray}
\usepackage{url}
\usepackage{caption}
\usepackage{bm}
\usepackage{makecell}
\usepackage{wrapfig,lipsum}
\usepackage{subcaption}
\usepackage{booktabs}
\usepackage[capitalize]{cleveref}
\crefformat{section}{\S#2#1#3}
\crefformat{subsection}{\S#2#1#3}
\crefformat{subsubsection}{\S#2#1#3}

\usepackage{tikz}
\usepackage{tabularx}

\usetikzlibrary{arrows}

\newcommand{\method}[1]{Counterfactual Attentiveness Test}
\newcommand{\accro}[1]{\textsc{CAT}}
\newcommand{\metric}[1]{Counterfactual-input sensitivity}

\newcommand{\adj}[1]{attentiveness}
\newcommand{\adjverb}[1]{attentive}
\newcommand{\ndata}[1]{ten}

\newcommand{\CAT}{\raisebox{-2pt}{\includegraphics[width=0.15in]{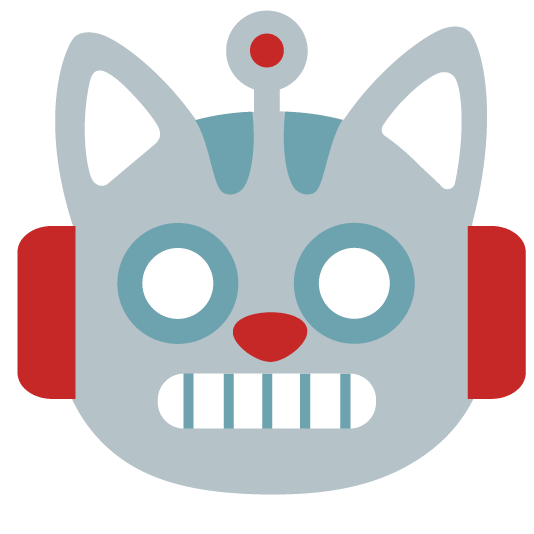}}}

\title{Measuring and Improving Attentiveness to Partial Inputs with Counterfactuals}

\author{Yanai Elazar\textsuperscript{1,2} \quad
Bhargavi Paranjape\textsuperscript{2}\Thanks{ Equal contribution.} \quad
Hao Peng\textsuperscript{3}$^*$ \quad
Sarah Wiegreffe\textsuperscript{1,2}$^*$ \\
{
\bf Khyathi Raghavi Chandu\textsuperscript{1} \quad
\bf Vivek Srikumar\textsuperscript{4} \quad
\bf Sameer Singh\textsuperscript{5} \quad
\bf Noah A.~Smith\textsuperscript{1,2}
}\\
\textsuperscript{1}Allen Institute for AI \\
\textsuperscript{2}University of Washington \\
\textsuperscript{3}University of Illinois Urbana-Champaign \\
\textsuperscript{4}University of Utah \\
\textsuperscript{5}University of California, Irvine \\
{\tt  yanaiela@gmail.com}\\
}

\begin{document}
\maketitle
\begin{abstract}

The inevitable appearance of spurious correlations in training datasets hurts the generalization of NLP models on unseen data.
Previous work has found that datasets with paired inputs are prone to correlations between a specific part of the input (e.g., the hypothesis in NLI) and the label; consequently, models trained only on those outperform chance.
Are these correlations picked up by models trained on the full input data?
To address this question, we propose a new evaluation method, \method{} (\accro{} \CAT{}).
\accro{} uses counterfactuals by replacing part of the input with its counterpart from a different example (subject to some restrictions), expecting an \textit{attentive} model to change its prediction. 
Using \accro{}, we systematically investigate established supervised and in-context learning models on \ndata{} datasets spanning four tasks: natural language inference, reading comprehension, paraphrase detection, and visual \& language reasoning. \accro{} reveals that reliance on such correlations is mainly data-dependent.
Surprisingly, we find that GPT3 becomes \textit{less} \adjverb{} with an increased number of demonstrations, while its accuracy on the test data improves.
Our results demonstrate that augmenting training or demonstration data with counterfactuals is effective in improving models' \adj{}. 
We show that models' \adj{} measured by \accro{} reveals different conclusions from solely measuring correlations in data.\footnote{Code available at: \url{https://github.com/yanaiela/partial_input}.}

\end{abstract}

\section{Introduction}

Reliance on spurious correlations compromises NLP models' abilities to generalize across different domains and tasks \citep{naik-etal-2018-stress,mccoy-etal-2019-right}. %
Intuitively, spurious correlations are features that are useful %
in the training data but are unreliable in general \citep{eisenstein-2022-informativeness}. 
Different approaches have been proposed for measuring model reliance on spurious correlations at various granularities, e.g., the appearance of certain tokens in a text, the length of a text, etc. \citep{sinha2021masked,gardner2021competency}.%

\begin{figure}[t!]
\centering
\includegraphics[width=.9\columnwidth]{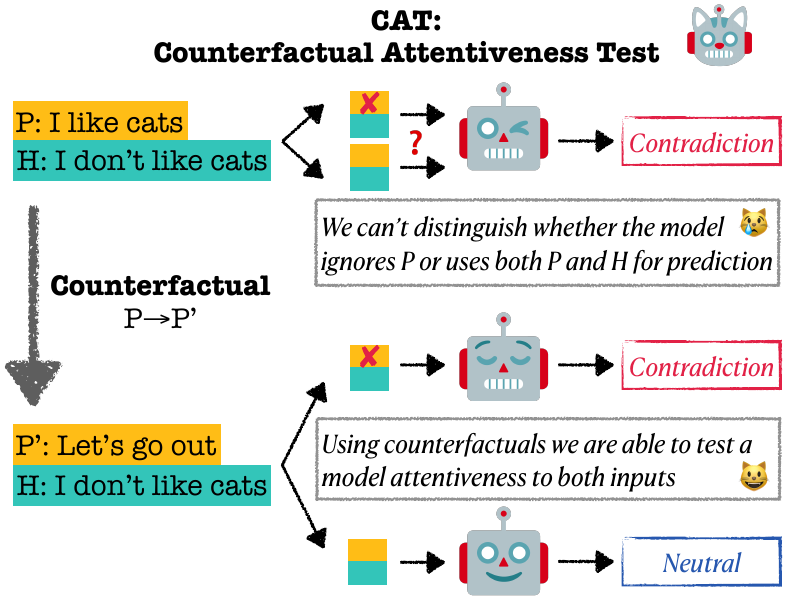}
\caption{We are interested in quantifying model's attentiveness to part of the input. 
We propose \textsc{\accro{}}, an evaluation that replaces the premise $P$ with a counterfactual $P'$. A change in behavior to the counterfactual input indicates the model is attentive to the premise, otherwise, the model relies solely on the hypothesis to make a prediction.}
\vspace{-1.7em}
\label{fig:intro}
\end{figure}

Spurious correlations in datasets are very common. 
For instance, 
the \textit{hypothesis-only baseline} \citep{poliak-etal-2018-hypothesis,gururangan-etal-2018-annotation} was designed to measure spurious correlations in natural language inference (NLI) datasets by training a model only on one part of the input (e.g., the hypothesis), effectively discarding information deemed critical for performing the task. 
Surprisingly, it performs significantly better than random guessing, despite never seeing the premise. 
These results suggest that  the hypothesis-only baseline has taken the shortcuts between the hypotheses and the labels during training.
However, they should not be interpreted as evidence that models trained on the full inputs (\textit{full-input models}) also take these shortcuts.
The question that then arises is \emph{how to quantify the reliance} of full-input models on the spurious correlations present in the hypotheses.

We propose a method for measuring models' reliance on partial input: \textsc{\method{}}, (\textsc{\accro{}}, \S\ref{sec:partial-input}). 
We present an overview of the approach in Figure \ref{fig:intro}. 
\accro{} is simple, intuitive, and is based on the idea of counterfactuals from the causal inference literature \citep{pearl}.
Importantly, it only requires a labeled dataset for the task and black box model predictions.
To test a model %
that performs a tasks with paired-inputs (tasks that consist of at least two components, e.g., NLI), we query the model, and replace one part of the input (e.g., the premise) with that part from another instance randomly drawn from the same data split. %
Intuitively, this perturbation results in the \textit{default} label for that task (typically a label that signifies no relation between the two inputs, e.g. \textit{neutral} in NLI).
As such, we focus on the subset of instances where the model predicts \textit{non-neutral}.\footnote{Other datasets and tasks may have different \textit{neutral} labels. We provide more details on these labels in \S\ref{sec:partial-input}.}
If the prediction changes, it suggests that the model is \textit{\adjverb{}} to the perturbed input. Conversely, a model that keeps the prediction unchanged is likely to be \textit{not \adjverb{}}, as it insensitive to the counterfactual. 
Following this intuition, our metric measures \textit{\adj{}} by calculating the percentage of predictions that \textit{changed} from the non-neutral labels on the original instances.

We conduct extensive experiments on four different tasks, \ndata{} datasets, and 15 different models, using two setups: supervised and in-context learning (\S \ref{sec:setup}).
We first extend previous work to obtain new results using the \textit{partial input baseline}---a model trained only on partial inputs, 
that quantifies spurious correlations in the data. %
Then, using our \adj{} metric, we show that even in datasets with spurious correlations between parts of the inputs and the labels, models do not always rely on them (\S \ref{sec:results}).
Finally, we study whether data augmentation using our counterfactuals improves models' \adj{} and find that is often the case (\S\ref{sec:counterfactual-training}).
Our results indicate that the appearance of --- often unavoidable --- spurious correlations in the data do not indicate models rely on them, and propose a simple and easy-to-use test for measuring it.

\section{Background \& Related Work}
\label{sec:related}
In this section we review the \textit{hypothesis-only} baseline in NLI and its extensions to other tasks: the \textit{partial input baseline}.
We then discuss a related work by \citet{srikanth-rudinger-2022-partial} and highlight the similarities and differences to our work.

\paragraph{Partial Input Correlations}

The hypothesis-only baseline refers to a supervised classifier trained only on the hypotheses in an NLI dataset, removing the premises from the original paired inputs \citep{gururangan-etal-2018-annotation,poliak-etal-2018-hypothesis,tsuchiya2018performance}. 
Predicting an NLI label on a single-sentence input is ill-defined,
but nonetheless, classifiers trained solely on the hypotheses on datasets such as SNLI \citep{bowman2015large} are able to generalize and achieve better-than-random accuracy on the corresponding test sets. This result indicates that the dataset used to train the hypothesis-only classifier contains predictive information about the labels. This phenomenon is often referred to as \textit{spurious correlations} because, in principle, both the premise and hypothesis should be required to infer the label. 
For instance, \citet{poliak-etal-2018-hypothesis} found that words such as `no', `sleeping', and `cat' were often used for generating hypotheses that contradict the premises, while `instrument', and `touching' were used more with entailed hypotheses. 
Other works discovered similar behavior in other tasks and datasets \citep{kaushik2018much,DBLP:conf/eccv/KVM20,trivedi2020multihop,mihaylov2018can,hessel2020does}.

\paragraph{NLI Models Reliance on the Hypothesis}

Recently, \citet{srikanth-rudinger-2022-partial} raised the question of whether models trained on the \emph{full input} data instances are in fact using such correlations (e.g., in NLI, ignoring the premise when making a prediction). They proposed to use counterfactuals to study whether full-input models rely on such heuristics. Importantly, \citet{srikanth-rudinger-2022-partial} manually constructed the counterfactuals, which limits the scale of such analysis. Their study uses RoBERTa-base and two NLI datasets (SNLI \citep{bowman2015large} and $\delta$-NLI \citep{rudinger2020thinking}) which led to the conclusion that models are \emph{not} relying on such heuristics. %
Our study covers a larger scope, including more tasks, datasets and setups (we also consider in-context learning), and reaches a more nuanced conclusion. We find that models trained on full inputs may still rely on such heuristic, while it is task and data-dependent.
We provide a detailed comparison with \citet{srikanth-rudinger-2022-partial}, and summarize the similarities and differences in Appendix \ref{sec:comparison}.

\paragraph{Counterfactuals for Evaluating NLP Models}
Counterfactuals were studied before in NLP. 
Some works use counterfactuals to study models' robustness to counterfactual perturbations \citep{glockner2018breaking,kaushiklearning,dev2020measuring,gardner-etal-2020-evaluating,wu2021polyjuice,fryer2022flexible,teney2020learning}. However, these works typically construct the counterfactuals manually, that involves expensive manual labor, in contrast to our method that produces counterfactuals automatically.
Finally, there is a line of research that modifies neural networks' parameters or intermediate states, to study high-level concepts, which are harder to create counterfactuals for, in the input-level \citep{serrano2019attention,elazar-etal-2021-amnesic,feder-etal-2022-causal,wu2022causal,geiger2021causal,geiger2022inducing}.

\section{Quantifying Attentiveness}
\label{sec:partial-input}

\begin{table*}[ht!]
\centering
\begin{adjustbox}{max width=0.9\textwidth}
\begin{tabular}{@{} l lll @{}}
    \toprule
    \textbf{Method} & \textbf{Train} & \textbf{Eval.} & \textbf{Eval. Ex} \\
    \midrule

    Full input & $\mathcal{M}_{x \rightarrow y}(x)$ & $\mathcal{M}_{x \rightarrow y}(x)$ & P: The other men shuffled. H: The other men were shuffled around. \\
    Prior work (\S\ref{sec:partial-input-baseline}) &  $\mathcal{M}_{x_2 \rightarrow y}(x_2)$ & $\mathcal{M}_{x_2 \rightarrow y}(x_2)$ & H: The other men were shuffled around.\\
    Naive partial-inputs (\S\ref{sec:naive}) & $\mathcal{M}_{x \rightarrow y}(x)$ & $\mathcal{M}_{x \rightarrow y}(x_2)$ & H: The other men were shuffled around. \\
    \textsc{\method{}} (\S\ref{ssec:our-method}) & $\mathcal{M}_{x \rightarrow y}(x)$ & $\mathcal{M}_{x \rightarrow y}(x'_1,x_2)$ & P: The dog barked. H: The other men were shuffled around. \\
    \midrule
    \multirow{2}{*}{\makecell[l]{Counterfactual \adj{} data augmentation (\S\ref{sec:counterfactual-training})}} & $\mathcal{M}_{(x_1,x_2) \rightarrow y}(x)$ & $\mathcal{M}_{(x_1,x_2) \rightarrow y}(x)$ & P: The other men shuffled. H: The other men were shuffled around. \\
    & $\mathcal{M}_{(x'_1,x_2) \rightarrow y}(x)$  & $\mathcal{M}_{(x'_1,x_2) \rightarrow y}(x)$ & P: The dog barked. H: The other men were shuffled around.\\
    \bottomrule
\end{tabular}
\end{adjustbox}
\caption{Comparing the methods we consider. Unlike prior work (``hypothesis-only baselines''), our proposed counterfactual method does not change the model. %
We additionally show with counterfactual training that training on examples of counterfactual partial-inputs can improve performance on them. $\mathcal{M}$ refers to a model, $(x)$ is a paired input that consist of the two variables $(x_1,x_2)$, similarly to $x'_2: (x'_1,x'_2)$, and $y$ is the model's prediction.
}
\vspace{-0.5em}
\label{tab:overview}
\end{table*}

We seek a method that measures model attentiveness to a part of the input in paired-input tasks, such as the premise or hypothesis in NLI. 
Formally, assume an input $x$ contains two parts: $x: (x_1, x_2)$,
and let $\mathcal{M}_{x \rightarrow y}$ be a model trained to predict $y$ from $x$. 
We are interested in measuring the causal effect of $x_1$ on model $\mathcal{M}$. If such effect is prominent, we conclude that the model relies on $x_1$ for inference,\footnote{We leave it to future work to explore problems where the input has even more structure (i.e., more than two parts), but conceptually the generalization is straightforward.} otherwise it ignores it. 
By measuring such effect on a dataset we quantify the extent by which the model relies on $x_1$.
Cases where such effect is large are especially interesting when the underlying training data contains correlations between $x_2$ and $y$, since the model does not rely on such correlations during training.
We describe the \textit{partial input baseline} and a naive approach for measuring \adj{} and explain why they do not fit as such metric. We then describe our method, \accro{}, in \S\ref{ssec:our-method}. We summarize these approaches in Table \ref{tab:overview}.

\subsection{Partial Input Baseline}\label{sec:partial-input-baseline}
As discussed in \S\ref{sec:related}, previous approaches such as the hypothesis-only baseline \citep{poliak-etal-2018-hypothesis} train a model solely on part of the input, $x_2$ (e.g., the hypotheses in NLI and paragraphs in RC, $\mathcal{M}_{x_2 \rightarrow y}(x_2)$). %
However, this approach does not tell us whether a \textit{full-input} model ($\mathcal{M}_{x \rightarrow y}(x)$) also relies on parts of the inputs as it is a different model; it only tells us whether parts of the inputs contain predictive information about the label. 
Therefore, the partial input baseline solely discovers spurious correlations in a dataset, and is not suitable for studying full-input models' behavior.

\subsection{Naive Solution}\label{sec:naive}
Instead of relying on a partial input model, we first introduce another naive approach to quantify a model's reliance on $x_2$, and explain why it is not valid for answering our research question.
Naively, one may use a full-input model $\mathcal{M}_{x \rightarrow y}(x)$ and evaluate the model on a partial input $\mathcal{M}_{x \rightarrow y}(x_2)$.
Then, we would test whether the predictions are equal, which would indicate that the model only relies on $x_2$.
However, attributing \adj{} (or inattentiveness) is problematic since we cannot disentangle the cause of such behavior, which can result from either the model's \adj{}, or from the model's behavior on out-of-domain data distribution \citep{fong2017interpretable,hooker-interpretability-benchmark}.

\subsection{\method{}}\label{ssec:our-method}

We propose an alternative solution that can be computed automatically, and provides a better estimate of the model's reliance on parts of the input. We call this method \textsc{\method{}} (\textsc{\accro{}} \CAT{}).
The method is inspired by \emph{counterfactuals} from causal inference to ask ``what would happen if part of the input was different?'' \citep{pearl}.
Instead of completely removing a part of the input, e.g., $x_1$, we replace it with $x'_1$ and evaluate the model's prediction on $\mathcal{M}_{x \rightarrow y}(x'_1,x_2)$.
Intuitively, replacing $x_1$ by $x'_1$ is likely to result in a non-related pair. As such, in almost all cases\footnote{We confirm this assumption empirically in \S\ref{subsec:verification}.} the new label is \textit{neutral} in the case of NLI.\footnote{Each task and dataset have their own corresponding labels. We discuss those labels in \S\ref{sec:setup}.}

As we are interested in measuring models' \adj{} with a change in prediction, we cannot estimate it in cases where the model predicts neutral on the original $(x_1,x_2)$ pair. We thus only consider cases where the prediction on the original pair is non-neutral.
Since the predictions on the original $(x_1,x_2)$ pairs may differ between models, we cannot compare them directly. 
However, in Appendix \S \ref{app:comparable}, we compute the \adj{} on the subset where models' predictions are identical and show that since the original predictions are often similar, the trends are consistent.

\paragraph{Attentiveness Metric}
We use accuracy to measure the \adj{} of a model on the counterfactual instances. %
Specifically, we consider as an initial set all of the instances $(x_1,x_2)$ from the development set.
We obtain the model's prediction and only keep the instances where the model predicts a non-\textit{neutral} label. 
Then, we generate a counterfactual by randomly sampling an instance $(x'_1, x'_2)$  from the same data split, and combining $x_2$ with $x'_1$.\footnote{As opposed to previous work \citep{srikanth-rudinger-2022-partial} that relies on hand-crafted counterfactuals to probe the model, we rely on much easier-to-collect counterfactuals, which can be randomly sampled from the data and do not require manual curation.}
Finally we compute how often the model changes its prediction from the initial pair on this subset. A higher score indicates the model is more \adjverb{} to the full input, which signifies the model does not rely on partial inputs for making predictions.
For each instance $(x_1,x_2)$ in the dataset, we sample $k=5$ counterfactuals $x'_1$ from the data and report the mean and standard deviation accuracy across the samples.

\begin{table*}[ht!]
\centering
\begin{adjustbox}{max width=1.\textwidth}
\begin{tabular}{@{} llllll @{}}
    \toprule
    \textbf{Task} & \textbf{Dataset} & \multicolumn{3}{c}{\textbf{Counterfactual Example}} & \textbf{Acc} \\
    \midrule
    & & \textbf{$x_1$} & \textbf{$x_2$} & \textbf{$y$} & \\
    \midrule
    \multirow{7}{*}{NLI}
    & MNLI & \makecell[l]{Perhaps North Africans and eastern Europeans peopled \\ the Ligurian coast, while the Adriatic and south may have been \\ settled by people from the Balkans and Asia Minor.} & \makecell[l]{You can find more information from \\ the senior executive's plan.} & neutral & 100\\
    \cline{2-6}
    & WANLI & \makecell[l]{The best known is the building that houses the National Gallery \\ in Trafalgar Square, London, designed by Sir Charles Barry and  \\ completed in 1843.} & \makecell[l]{A great many of the villages in the area \\ are in a state of repair, and some of them \\ are actually inhabited.} & neutral & 100 \\
    \cline{2-6}
    & RTE & \makecell[l]{Schroder Investment Management has indicated its intention to \\ accept Revival's offer to buy retailer Marks \& Spencer.} & \makecell[l]{There are 32 pandas in the wild in China.} & non-entailment & 100 \\
    \midrule
    \multirow{2}{*}{PD} 
    & QQP & \makecell[l]{How do you delete messages on Snapchat?} & \makecell[l]{Which are some of the best companies \\ to work for?} & non-paraphrase & 100 \\
    \cline{2-6}
    & PAWS & \makecell[l]{The Arieşul Mare River is a tributary of the Vâlcea River in Romania.} & \makecell[l]{Also steam can be used and need not be pumped.} & non-paraphrase & 100 \\
    \midrule
    \multirow{9}{*}{RC}    
    & SQuAD 2.0 & \makecell[l]{Hypersensitivity is an immune response that damages \\ the body's own tissues.
    They are divided into four classes\\ (Type I – IV) based on the mechanisms involved\dots\\ These reactions are mediated by T cells, monocytes, and macrophages.} & \makecell[l]{What tracts does commensal flora \\ help pathogens thrive in?} & no answer & 98 \\ %
    \cline{2-6}
    & DuoRC & \makecell[l]{ South Boston teenager Jason Tripitikas is a fan of martial arts films and \\ awakens from a dream of a battle between the Monkey King and celestial \\soldiers in the clouds. He visits a pawn shop in Chinatown to buy \\ Wuxia DVDs and discovers a golden staff. On his way home, Tripitikas} &  What is Ana's profession? & no answer & 100 \\ %
    \cline{2-6}
    & NewsQA & \makecell[l]{Iran's parliament speaker has criticized U.S. President-elect Barack Obama \\ for saying that Iran's development of a nuclear weapon is unacceptable. \\ Iranian President Mahmoud Ahmadinejad has outlined where  he thinks  U.S. \\ policy needs to change. Ali Larijani said Saturday that Obama should apply \\ his campaign message of change to U.S. dealings with Iran.} & What does the U.N drug chief advocate? & no answer & 98 \\ %
    \midrule
    \multirow{6}{*}{VLR} & VQA 2.0 & \includegraphics[width=0.6in, valign=m]{figures/dataset_examples/COCO_val2014_000000016451.jpg} & What color are this person's shoes? & no answer
    & 92 \\
    \cline{2-6}
    & \makecell[tl]{NLVR2}
    & 
    \includegraphics[width=0.7in, valign=m]{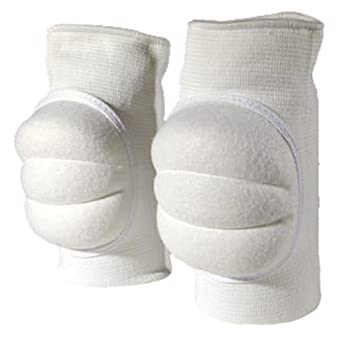}
    \includegraphics[width=0.7in, valign=m]{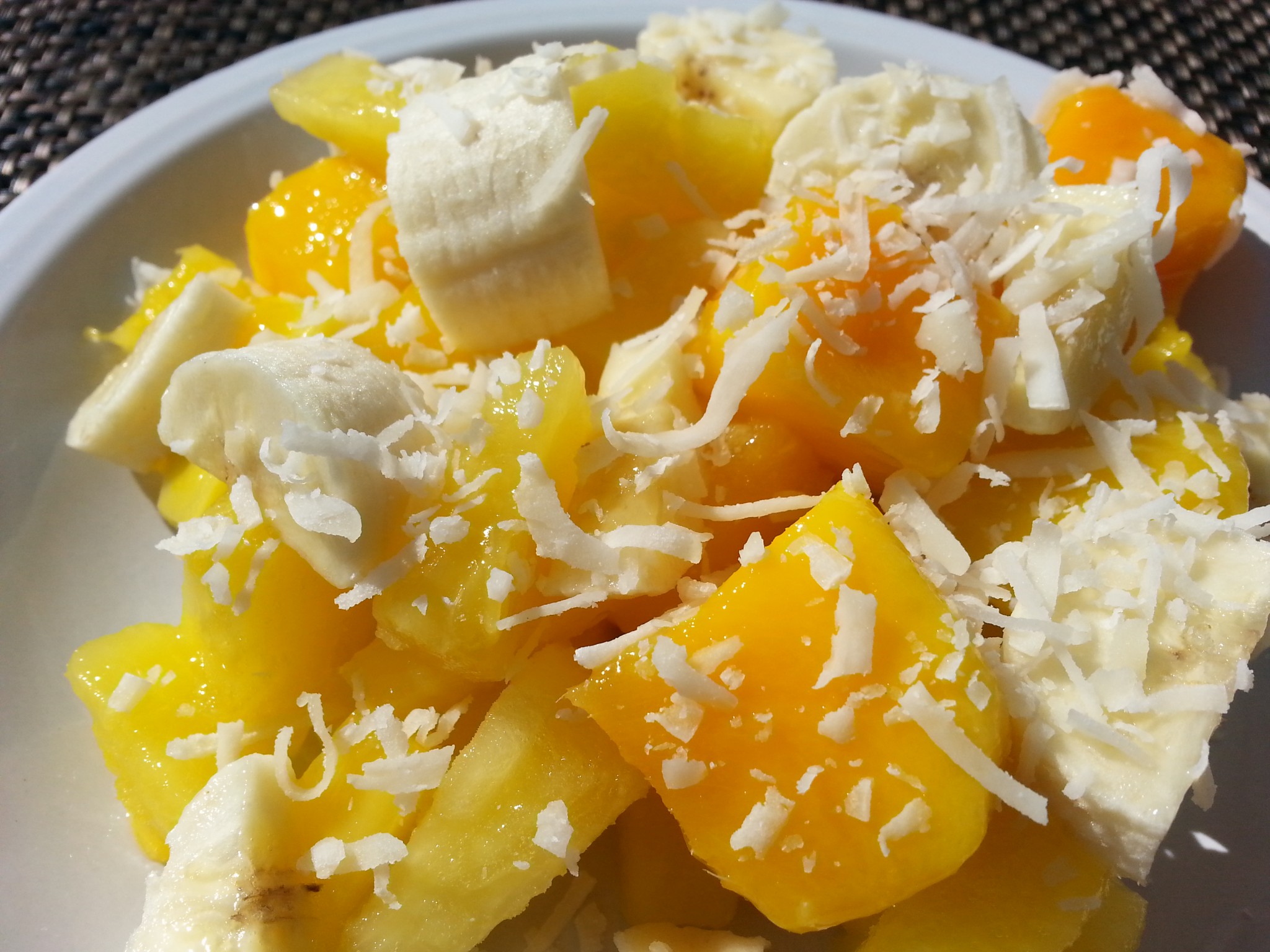} & There is a smartphone in the right image. & false & 96 \\
    \bottomrule
\end{tabular}
\end{adjustbox}
\caption{An example of a counterfactual pair $(x'_1,x_2)$ from each of the datasets we consider. The label $y$ is assigned automatically to the counterfactual data. ``Acc.'' indicates the \% of instances from our manual annotation experiment (\cref{subsec:verification}) where the assumption holds that the model's predicted label should change from the original to the counterfactual instance if the model is attentive.
}
\label{tab:examples}
\end{table*}

\section{Experimental Setup}
\label{sec:setup}

\begin{table}[t!]
\centering
\begin{adjustbox}{max width=.9\columnwidth}
\begin{tabular}{@{} clll @{}}
\toprule

\bf Setup & \bf Task & \bf Dataset &\bf Model \\
\midrule

\multirow{4}{*}{\rotatebox[origin=c]{90}{Supervised}}
    & NLI & RTE, MNLI, WANLI & \multirow{3}{*}{\makecell[l]{BERT, RoBERTa, \\ DeBERTa, T5, \\ T5-v1.1, Flan-T5}} \\
    & PD & QQP, PAWS & \\
    & RC & SQuAD2.0, DuoRC, NewsQA & \\ \cmidrule{2-4}%
    & VLR & VQA2.0, NLVR2 & BLIP, ViLT \\
    \midrule
\multirow{2}{*}{\rotatebox[origin=c]{90}{ICL}}
    & NLI & RTE, MNLI, WANLI & \multirow{2}{*}{GPT-3, T5, Flan-T5} \\
    & PD  & QQP, PAWS &  \\

\bottomrule
\end{tabular}

\end{adjustbox}
\caption{
Details about the configurations we use.
We experiment with ten datasets and 11 model families.}
\label{tab:datasets-models}
\end{table}

We consider four diverse tasks and \ndata{} English datasets. We also consider two learning strategies: supervised learning through fine-tuning using 12 models from six model families, and in-context learning (ICL) using nine models from three model families.
Table \ref{tab:datasets-models} summarizes all configurations.

\subsection{Tasks and Datasets}
We use the standard train-development splits\footnote{Except for WANLI which does not have a development set, for which we randomly sample 5K instances from the training set and use them for evaluation.} for each dataset. In the ICL setup, we randomly sample $k$ instances from the training sets.

\paragraph{Natural Language Inference} We consider three datasets for NLI: MNLI \citep{williams-etal-2018-broad}, WANLI \citep{liu-etal-2022-wanli}, and RTE \citep{dagan2005pascal}. MNLI and WANLI contain three labels: \textit{entailment}, \textit{neutral}, and \textit{contradiction}, while RTE only contains \textit{entailment} and \textit{non-entailment}.

\paragraph{Paraphrase Detection}
We consider two datasets for paraphrase detection (PD): Quora Question Pairs (QQP; \citealp{Sharma2019NaturalLU}) and PAWS \citep{zhang-etal-2019-paws}. These datasets contain questions that are labeled as \textit{paraphrase} or \textit{non-paraphrase}.

\paragraph{Reading Comprehension}
We consider three datasets for reading comprehension (RC) that include a subset of questions that are not answerable i.e., contain a \textit{no-answer} label: SQuAD2.0 \citep{rajpurkar-etal-2018-know}, DuoRC \citep{saha2018duorc}, and NewsQA \citep{trischler2017newsqa}.

\paragraph{Visual \& Language Reasoning}
We consider two datasets for visual and language reasoning tasks (VLR). VQA2.0 \citep{DBLP:journals/ijcv/GoyalKASBP19} contains questions and images with multiple choice answers. 
NLVR2 \citep{DBLP:conf/acl/SuhrZZZBA19} consists of two images and a statement where the task is to predict whether the statement is true for the images. 

\subsection{Counterfactual Default Labels}
\accro{} generates counterfactuals by automatically sampling two inputs from different instances. This makes an assumption that no relation would hold between the inputs in the new counterfactual instance (as shown in the examples in Table \ref{tab:examples}), together with the respective label.
Such relations are different between tasks and datasets, which we now specify:
\textit{neutral} in 3-class NLI (e.g. MNLI), \textit{non-entailment} in 2-class NLI (e.g. RTE), \textit{not-paraphrase} in the PD, \textit{no-answer} in RC, \textit{false} in NLVR2, and \textit{no-answer} in VQA. 

\paragraph{Label-Flip Assumption Verification}
\label{subsec:verification}

We verify \accro{}'s assumptions by hand-annotating 50 counterfactually-paired instances per dataset for each of our four tasks. We annotate whether a representative model's predicted label for an instance is incorrect for the counterfactual instance and should thus change if the model is attentive. Overall we find the assumption to be true most of the time:
100\% of instances in MNLI, WANLI, RTE, QQP, PAWS and DuoRC; 98\% in SQuAD2.0 and NewsQA, 96\% in NLVR2,
and 92\% in VQA 2.0.\footnote{Cases where the assumption fails for VQA are primarily yes/no questions where the answer is still no after the image is shuffled (4/25). When the questions are shuffled, the assumption holds 100\% of the time (25/25).}
We provide some randomly-selected instances from each dataset in Table \ref{tab:examples}, along with the accuracy from our manual evaluation. 
 
\subsection{Models}

\paragraph{Supervised Learning}
We experiment with six text-only models and two multimodal models which were trained on some pretraining corpora, and then fine-tuned on a respective supervised dataset (e.g., MNLI). We use three masked language models: BERT \citep{devlin-etal-2019-bert}, RoBERTa \citep{Liu2019RoBERTaAR}, DeBERTa v3 \citep{he2021debertav3}. and three T5 models \citep{raffel2020exploring}, with both the v1.0, and v1.1 versions (v1.0 was first pre-trained, and then fine-tuned on supervised training data from GLUE \citep{wang-etal-2018-glue}, and QA tasks, while v1.1 was only pre-trained as a language model). Additionally, we also use Flan-T5 \citep{flant5}, T5 models that were fine-tuned on 1,836 tasks with instructions.
We use the base and large versions of each model. 
For multimodal models, we experiment with BLIP \citep{DBLP:conf/icml/0001LXH22}, and ViLT \citep{DBLP:conf/icml/KimSK21}.

\paragraph{In-Context Learning}
We experiment with three types of model families that were shown to perform well in zero-shot and few-shot settings using in-context learning (ICL).
The first, text-davinci-003, is a GPT-3 model that was first trained on code and then was additionally trained to follow instructions \citep{ouyang2022training}. We refer to this model simply as GPT-3. Since the model is a commercial product, we do not have full information of its training data, and other crucial reproducibility details. For instance, it is possible that the evaluation data we test the model on were seen by the model. However, it is reasonable to assume that the model has not seen the paired input from \accro{}, as these are randomly constructed pairs, and those results can be trusted.
In addition, we experiment with two open-sourced models: T5 \citep{raffel2020exploring}, and Flan-T5 \citep{flant5}. Flan-T5 was initialized from the T5 model, and finetuned on 1,836 supervised tasks accompanied by instructions.

\subsection{Learning Strategies}

\paragraph{Supervised Learning through Finetuning}
We consider the fine-tuning setup to train a model per task and dataset. %
Each model in these experiments is trained for 10 epochs, following \citet{bandel-etal-2022-lexical} to reduce reliance on lexical overlap heuristics. We use the same learning rate for all models.\footnote{Except for the smaller models (T5-small and base) as they were unable to learn using smaller learning rates.} %

\paragraph{In-Context Learning}
We follow the prompt format in \citet{raffel2020exploring} for the T5 models. 
For GPT-3 and Flan-T5, we prepend instructions to each instance (described in Table~\ref{tab:icl_prompts} in the Appendix).\footnote{Preliminary experiments find that including instructions substantially improves both the accuracy and \adj{}.}
We consider both zero-shot and few-shot settings.
Each demonstration includes examples from all labels, and we consider $k \in \{0,1,2,3\}$ number of tuples, which translates into $\{0,2,4,6\}$ or $\{0,3,6,9\}$ depending on the number of labels in a dataset. We use greedy decoding for generating the answers for each of these models.

\section{Results}
\label{sec:results}

\paragraph{Standard Evaluation}\label{ssec:eval}
We begin by training models on their corresponding datasets. %
For example, DeBERTa-large achieves 87.73\%, 91.28\%, and 76.72\% accuracy on RTE, MNLI, and WANLI, respectively.\footnote{Our MNLI results are comparable to those by \citep{he2021debertav3},
but on RTE ours is worse by 5\%. This is likely because we do not perform extensive hyper-parameter search. Note that we achieve a reasonable performance nonetheless.}
The full results are in Appendix \ref{app:standard-acc}.

\paragraph{Partial Input Baseline}
\label{subsec:partial-input-baseline}
We experiment with the partial input baseline (\cref{sec:partial-input-baseline}) and train with only parts of the inputs. We extend previous work \citep{gururangan-etal-2018-annotation,kaushik-lipton-2018-much} and perform extensive experiments with different types of models, tasks, and datasets. Every experiment corresponds to the evaluation performance of a model trained on the respective training set from the same data.
The results are in Appendix \ref{app:partial-input-baseline}.

Hypothesis-only MNLI models achieve as high as 61.66\% accuracy with T5-1.1-large, a much higher accuracy than the majority baseline (35.4\%). On the other hand, models trained on RTE and WANLI only achieve 10.5\% and 7.4\% points better than majority.
For QA, the strong performance of the question-only models is particularly noteworthy. The best results with DeBERTa-large is 98.5\% on SQuAD2.0, 73.2\% on DuoRC, and 96.8\% on NewsQA. These models are pre-trained on a large amount of text data in the same domain as the respective datasets and are likely to select the correct answer span from a random passage based on their parametric knowledge and grammar heuristics. 

Our results support and extend prior findings by showing that for a diverse class of models and a variety of tasks, models trained on partial inputs are strong. Next, we revisit the idea that these results indicate standard models trained on such datasets also relies on such heuristics.

\subsection{\method{}}
\label{sec:counterfactuals}\label{ssec:cat}

\begin{table}[t!]
\centering
\begin{adjustbox}{max width=1\columnwidth}
\centering
\begin{tabular}{@{} lcc  m{0.01em}  c  m{0.01em}  c @{}}
\toprule
 & \multicolumn{2}{c}{\bf  NLI} && \multicolumn{1}{c}{ \bf  PD} && \multicolumn{1}{c}{ \bf RC}\\
\cline{2-3}
\cline{5-5}
\cline{7-7}
{\bf Dataset} & \bf RTE & \bf  MNLI && \bf  PAWS && \bf  SQuAD2.0 \\
\midrule

BERT-base & 99.8$_{ \pm 0.3}$ & 74.6$_{ \pm 0.3}$ &  & 99.3$_{ \pm 0.2}$ &  & 98.0$_{ \pm 0.0}$ \\
BERT-large & 100.0$_{ \pm 0.0}$ & 64.3$_{ \pm 0.1}$ &  & 85.9$_{ \pm 0.3}$ &  & 98.0$_{ \pm 0.2}$ \\
RoBERTa-base & 99.8$_{ \pm 0.3}$ & 73.4$_{ \pm 0.5}$ &  & 80.5$_{ \pm 0.6}$ &  & 98.3$_{ \pm 0.1}$ \\
RoBERTa-large & 100.0$_{ \pm 0.0}$ & 69.8$_{ \pm 0.2}$ &  & 78.7$_{ \pm 0.3}$ &  & 98.2$_{ \pm 0.2}$ \\
DeBERTa-base & 97.0$_{ \pm 1.6}$ & 78.8$_{ \pm 0.4}$ &  & 99.9$_{ \pm 0.1}$ &  & 97.6$_{ \pm 0.1}$ \\
DeBERTa-large & 100.0$_{ \pm 0.0}$ & 68.6$_{ \pm 0.3}$ &  & 99.1$_{ \pm 0.1}$ &  & 98.0$_{ \pm 0.2}$ \\
T5-base & 99.4$_{ \pm 0.7}$ & 81.3$_{ \pm 0.1}$ &  & 99.0$_{ \pm 0.1}$ &  & 99.1$_{ \pm 0.1}$ \\
T5-large & 100.0$_{ \pm 0.0}$ & 71.5$_{ \pm 0.4}$ &  & 100.0$_{ \pm 0.0}$ &  & 99.5$_{ \pm 0.1}$ \\
T5-1.1-base & - & 46.9$_{ \pm 0.6}$ &  & - &  & 99.4$_{ \pm 0.0}$ \\
T5-1.1-large & - & 79.4$_{ \pm 0.3}$ &  & - &  & 99.4$_{ \pm 0.1}$ \\
FLAN-T5-base & 99.9$_{ \pm 0.3}$ & 81.8$_{ \pm 0.3}$ &  & 98.3$_{ \pm 0.2}$ &  & 99.4$_{ \pm 0.1}$ \\
FLAN-T5-large & 100.0$_{ \pm 0.0}$ & 77.1$_{ \pm 0.2}$ &  & 100.0$_{ \pm 0.1}$ &  & 99.5$_{ \pm 0.1}$ \\

\bottomrule
\end{tabular}

\end{adjustbox}
\caption{\accro{} results on natural language inference (NLI), paraphrase detection (PD) and reading comprehension (RC). Results are calculated by computing the mean attentiveness of five counterfactuals per instance, and their standard deviation. Higher numbers indicate better model attentiveness. 
We mark results for models whose standard performance is not significantly better than random with `-'.
}
\vspace{-.8em}
\label{tab:counterfactuals-dev-main}
\end{table}

\begin{table}[t!]
\centering
\begin{adjustbox}{max width=1\columnwidth}

\begin{tabular}{@{} lrr @{} }
\toprule
\bf Dataset & \bf VQA v2 & \bf NLVR2 \\
\midrule
BLIP (image) & 65.1$_{\pm 0.0}$ & 95.9$_{\pm 0.3}$ \\
BLIP (text) & 89.6$_{\pm 0.0}$ & 96.0$_{\pm 0.4}$ \\
\midrule
ViLT (image) & 63.6$_{\pm 0.0}$ & 93.3$_{\pm 0.3}$\\
ViLT (text) & 89.3$_{\pm 0.0}$ & 93.1$_{\pm 0.3}$\\
\bottomrule
\end{tabular}

\end{adjustbox}
\caption{\accro{} results on visual reasoning (VLR). Results are calculated by computing the mean attentiveness of five counterfactuals per instance, and their standard deviation. Higher numbers indicate better model attentiveness. 
The portion of the input that is perturbed is indicated in parentheses.
}
\label{tab:vr-counterfactuals}
\end{table}

Next, we report the results of our \method{} on the different setups. 
Overall, the main trend we observe in our experiments is that \adj{} is data and setup specific.
The results are summarized in Table \ref{tab:counterfactuals-dev-main} for the supervised textual tasks, Table \ref{tab:vr-counterfactuals} for the visual reasoning tasks, and Table \ref{tab:icl_counterfactual} for ICL classification tasks.\footnote{The full results are presented in the Appendix (Table \ref{tab:counterfactuals-dev}).} %

\paragraph{Supervised Models}
Attentiveness scores are consistent across models, e.g., on RTE all models are highly attentive with average scores ranging from 97.0\% to 100\%. On the other hand, models trained on MNLI are much less attentive, with scores ranging from 46.9\% (T5-1.1-large) to 81.8\% (FLAN-T5-base).
In the case of VLR, we test the attentiveness of models to both parts of the input, once to the image and once to the text. The results indicate that models are more attentive to the images, than to the texts. For instance, in VQA v2, when perturbing the images, BLIP achieves 65.1\%, but when perturbing the texts it gets 89.6\%.

\paragraph{ICL}
We report the ICL attentiveness results in Table \ref{tab:icl_counterfactual}.
The results in this setup follow a similar trend to the supervised models, where attentiveness results are data dependent. On RTE, models are highly attentive, performing 100\% on CAT in most cases, and above 98\% in the other cases. On MNLI however, the attentiveness varies from 28.3\% (T5-3B, one-shot), 62.5\% (GPT3, 3-shot) to 98.9\% (Flan-T5-XL, 0-shot).

\subsection{Partial Input Baseline as Indication of Attentiveness?}

\begin{figure}[t!]
    \centering
    \includegraphics[width=1\columnwidth]{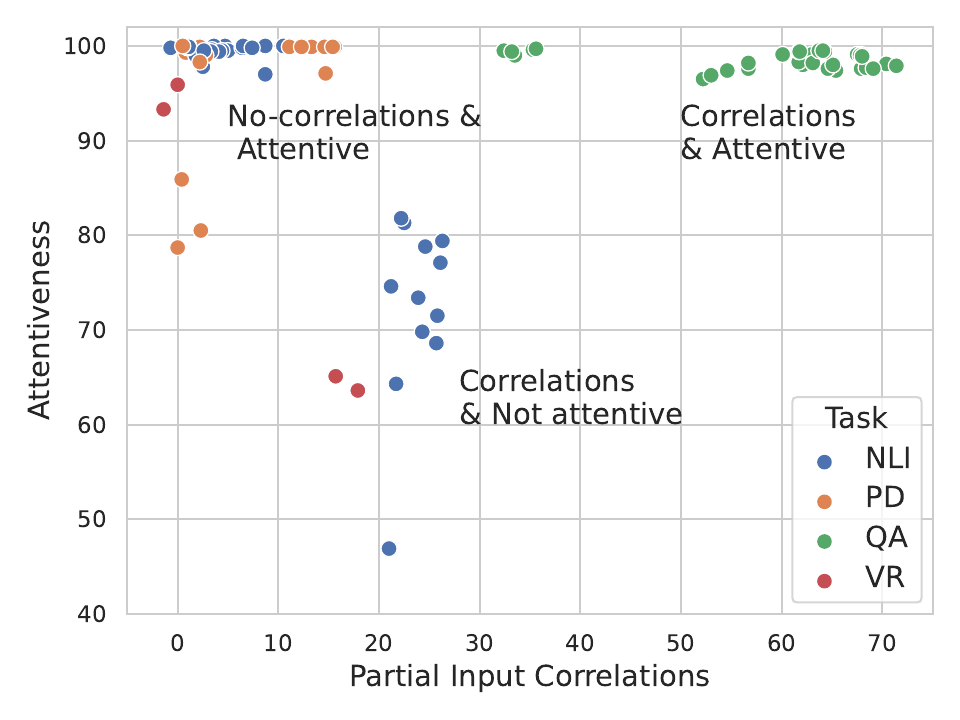}
    \caption{Attentiveness as a function of partial input correlations for the supervised models. Higher attentiveness values ($y$ axis) indicate better attentiveness to counterfactual inputs. Higher values on the partial input correlations indicate correlations between a part of the input and the label. It is computed as the score on the \textit{partial input} baseline, subtracting the majority score.
    }
    \label{fig:robustness}
    \end{figure}

Next, we study the relationship between the \textit{partial input} baseline performance as a method to estimate the spurious correlations in a dataset, and the attentiveness of a model and its ability to overcome these correlations. This relationship is presented in Figure \ref{fig:robustness}. We plot the results of the supervised models, trained on the different considered datasets. 
On the x-axis, we plot the difference between the trained partial input baseline model and random accuracy (\textit{partial input correlations}), and on the y-axis, the model's attentiveness score.

First, we highlight the left part of the figure where the spurious correlations are small: models trained on datasets with less than a 10\% gap with the partial input baseline over the majority label are mostly robust, with attentiveness scores as high as 100\% (with 3 outliers, where the smallest one achieves 78.7\%). This is unsurprising, as low partial input correlations indicate fewer spurious correlations in the dataset, so full-input models should be able to detect counterfactual inputs more readily.
Next, we observe scattered patterns, typically clustered based on tasks and datasets, and overall worse attentiveness. For instance, the middle cluster consists of NLI models trained on MNLI, with high partial input correlations 
(between 20-30\%) and low attentiveness (between 56.9-81.8\%).
Interestingly, models trained on RC datasets appear to be attentive (above 95\%), while also having high partial input correlation scores (32.4\%-71.4\%).
These results highlight two findings:
(1) low partial input correlations in the training data result in high attentiveness, and (2) \textbf{high partial input correlations are not a good indicator of standard model behavior}: some models (trained on certain datasets) rely on such heuristics, whereas others do not.

\begin{table}[t!]
\centering
\begin{adjustbox}{max width=1.\columnwidth}
\begin{tabular}{@{} l c cc m{0.001em} c   @{}} %

\toprule
&& \multicolumn{2}{c}{\textbf{NLI}} && \multicolumn{1}{c}{\textbf{PD}}\\
\cline{3-4}
\cline{6-6}
\bf Model & \bf \#Tuples & \bf RTE & \bf MNLI  && \bf PAWS\\
\midrule

GPT3 & 0
& 99.5$_{\pm 0.4 }$
& 92.9$_{\pm 0.3 }$
&
& 99.9$_{\pm 0.0 }$
\\
GPT3 & 1
& 100.0$_{\pm 0.0 }$
& 66.0$_{\pm 0.3 }$
&
& 100.0$_{\pm 0.0 }$
\\
GPT3 & 2
& 100.0$_{\pm 0.0 }$
& 62.5$_{\pm 0.5 }$
&
& 100.0$_{\pm 0.0 }$
\\
GPT3 & 3
& 100.0$_{\pm 0.0 }$
& 62.0$_{\pm 0.3 }$
&
& 99.9$_{\pm 0.0 }$
\\

\midrule\midrule

Flan-T5-base & 0
& 100.0$_{\pm 0.0 }$
& 66.5$_{\pm 0.3 }$
&
& 99.9$_{\pm 0.0 }$
\\
Flan-T5-base & 1
& 98.1$_{\pm 1.4 }$
& 48.6$_{\pm 0.4 }$
&
& 100.0$_{\pm 0.0 }$

\\\midrule
Flan-T5-large & 0
& 100.0$_{\pm 0.0 }$
& 94.8$_{\pm 0.2 }$
&
& 100.0$_{\pm 0.0 }$
\\
Flan-T5-large & 1
& 99.7$_{\pm 0.3 }$
& 96.2$_{\pm 0.3 }$
&
& 100.0$_{\pm 0.0 }$

\\\midrule
Flan-T5-XL & 0
& 100.0$_{\pm 0.0 }$
& 98.9$_{\pm 0.1 }$
&
& 100.0$_{\pm 0.0 }$
\\
Flan-T5-XL & 1
& 100.0$_{\pm 0.0 }$
& 98.6$_{\pm 0.2 }$
&
& 100.0$_{\pm 0.0 }$
\\\midrule

T5-base & 0
& 100.0$_{\pm 0.0 }$
& 76.6$_{\pm 0.3 }$
&
& -
\\
T5-base & 1
& - %
& 51.4$_{\pm 3.4 }$
&
& -
\\\midrule
T5-large & 0
& 100.0$_{\pm 0.0 }$
& 72.9$_{\pm 0.5 }$
&
& -
\\
T5-large & 1
& 99.3$_{\pm 0.6 }$
& 43.5$_{\pm 5.5 }$
&
& -
\\\midrule
T5-3B & 0
& 100.0$_{\pm 0.0 }$
& 70.5$_{\pm 0.3 }$
&
& -
\\
T5-3B & 1
& - %
& 28.3$_{\pm 4.5 }$
&
& -
\\

\bottomrule
\end{tabular}

\end{adjustbox}
\caption{\accro{} performance using ICL. %
We report the mean and standard deviation across five counterfactuals per instance.
Models whose standard performance is not significantly better than random are marked with `-'.
}
\label{tab:icl_counterfactual}
\vspace{-3mm}
\end{table}

\subsection{ICL: The Role of Demonstrations}
The standard accuracy (Table \ref{tab:icl_acc} in the Appendix) shows two trends: (1) more demonstrations improve GPT's accuracy, but generally decrease the accuracy for others. Can we rely on these results as the main measure for model behaviors?
ICL attentiveness results (Table \ref{tab:icl_counterfactual}) tell a different story. While on RTE, and the two PD datasets the attentiveness performance stays the same - near perfect, the attentiveness on MNLI and WANLI decreases when more demonstrations are used.%

One hypothesis for GPT-3's low attentiveness is benchmark contamination \citep{rogers2023, jacovi2023stop}. If the model was exposed to the datasets we investigate and memorized their labels, this could result in high standard accuracy (which we observe). The low attentiveness we observe similarly supports the memorization hypothesis, since our perturbations are far less likely to have been observed by the model during training, so a model that has memorized rather than generalized would not do well on them.
These results highlight the importance of reporting metrics such as attentiveness when clean evaluation sets are not available. %

\section{\accro{} Augmentation}
\label{sec:counterfactual-training}

Considering the simplicity and automatic nature of \accro{}, we reuse the same sampling process for data augmentation to improve non \adjverb{} models. 
We follow a similar procedure as in the evaluation to augment the training data: apart from the regular training set, for every $x_1, x_2$ pair, where the label $y$ is different than the default label (e.g. \textit{neutral} for NLI), we sample $x'_1$ from the training data, assigned the default label.\footnote{This will increase the data size by the proportion of non-default labels, e.g. the new training set size will increase by 66\% in the case of balanced 3-label classification like MNLI.}

\begin{table}[t!]

\centering
\begin{adjustbox}{max width=1.\columnwidth}

\begin{tabular}{@{} llrr @{}}
\toprule
\bf Setup & \bf Model & \bf Acc. & \bf Attentiveness \\
\midrule
\multirow{2}{*}{Finetune} & T5-large & 89.3 & 71.5$_{\pm0.4}$ \\
& \quad +augmentation & 89.3 & \textbf{99.7$_{\pm0.0}$} \\
\cmidrule(lr){2-4}

\multirow{3}{*}{ICL} & GPT3, 4 tuples, half neutral 
& 78.1 
& 69.3$_{\pm0.2}$ \\ 

& GPT3, 3 tuples 
& 77.4 
& 66.2$_{\pm0.3}$ \\ %

& \quad +augmentation & 66.4 
& \textbf{76.7$_{\pm0.3}$} \\ %

\bottomrule
\end{tabular}

\end{adjustbox}
\caption{Standard accuracy performance on the respective development sets models are trained on, and the counterfactual accuracy of our metric.}
\vspace{-.5cm}
\label{tab:augmentation}
\end{table}

The following experiments study the models that perform poorly on our metric in the standard setup. 
We focus on MNLI since models are the least attentive on this dataset, and experiment once with T5-large in the supervised setup, and with GPT-3 in the ICL setup, using the original data (or demonstrations), and add the counterfactual instances.
The labels of counterfactuals match the default label - \textit{neutral} for MNLI.
We report both accuracy on the standard development set that verifies the augmentation does not degrade model performance and attentiveness. For each dataset, we report the performance of the vanilla model (those trained in \cref{ssec:eval} and used for \accro{} in \cref{ssec:cat}) and the augmented model. The results are in Table \ref{tab:augmentation}, which can be directly compared to the results of training on the standard datasets in \cref{tab:standard-perf,tab:icl_acc} in the Appendix.

For supervised learning, augmentation has little impact on standard accuracy but improves attentiveness. T5-large achieves 71.5\%, and 99.7 in attentiveness before and after the augmentation, respectively (and 89.3\% accuracy both before and after augmentation on the MNLI development set).

For GPT3, we expeirment with two baselines, one with 3 demonstration tuples with balanced labels, and the other with four, half of which having the neural label.
The latter aims to control for the label distribution shift invited by the data augmentation process.
The trend is different than the finetuend T5-large: GPT-3 with data augmentation sees decreased accuracy and increased attentiveness, but still far from perfect attentiveness.
These results suggest that this simple augmentation procedure can be used as a method for improving model's robustness, as measured by \accro{}.

\section{Conclusions}
\label{sec:conclusion}
We studied models' reliance on partial inputs and proposed a method to quantify it. 
We applied our method, \method{} (\accro{}) on multiple tasks, datasets, and setups including fine-tuning and in-context learning. 
We found that the spurious correlations between a part of the inputs and the labels are not always indicative of the full model's reliance on such heuristics in the supervised setup. 
For instance, we found that while partial inputs in SQuAD2.0 are correlated with the labels, models trained on it are attentive to both parts of the input.
This is in contrast to MNLI, whose hypotheses and labels are highly correlated, and models trained on it are also inattentive to the premises.
In addition, in ICL, we found that more demonstrations lead to decreased attentiveness in models such as GPT-3.

\section{Limitations}

While \accro{} allows automatic evaluation of the \adj{} of a model on the full input, it has some limitations.

\paragraph{Not a Standalone Evaluation} 
\accro{} is not meant to replace the standard model's performance, but to provide insights into models' behavior.
For instance, a model can perform close-to-perfect on \accro{} if the model relies on lexical overlap heuristics, as the counterfactual pairs are not likely to have high lexical overlap. As such, while the model is likely to perform well on \accro{}, it will fail on other benchmarks (e.g., HANS, \citealt{mccoy-etal-2019-right}).
In addition, \accro{} will not produce a meaningful score for models that did not learn the task and always predict the same label.
We highlight the importance of reporting both the standard metric used for a task or dataset, as well as ours for a more complete evaluation.

\paragraph{Random Counterfactual Assumption}

The automatic counterfactual generation  process assumes that two random instances from the same dataset will not be related, and as such the label will change. However, this is an assumption that should be manually evaluated. As we discuss in Section \ref{subsec:verification}, and in \autoref{tab:examples}, we show that for the datasets we considered this assumption is mostly true (with the smallest percentage being 92\% correct). 
However, in preliminary experiments we considered another dataset where this assumption didn't hold in many cases: SNLI \citep{bowman2015large}. Out of 50 counterfactual instances we manually inspected, we found the \textit{neutral} label to be correct only in 70\% of the cases. 
Consider the following counterfactual example we generated from SNLI: ``P: A man standing in front of a chalkboard points at a drawing. H: The person is in a hat has a big bag while walking on a tough terrain.''. We labeled this example as \textit{contradiction} because both sentences discuss the same entity. This is as opposed to the original \textit{entailment} label this hypothesis was paired with the following premise: ``P: A person in a red hat with a huge backpack going hiking.''.
The neutral label assumption does not hold in the SNLI case is mainly due to the source of data used to collect SNLI, which was based on image captions. In addition, the dataset lexicon is not diverse enough, and we often observe the same entities mentioned in the generated counterfactuals. Inspecting the SNLI gold annotations, it is clear that when an entity is mentioned in both P \& H, one must assume that they refer to the same entity if reasonably possible.\footnote{Some of the common mentioned entities are: man, woman, boy, people.}
This result led us to discard SNLI from the analysis, as our assumption was not accurate enough.

SNLI is the only additional dataset we considered for this study and did not include in the analysis due to our assumption not holding in practice.
We note though, that when analyzing other datasets, a practitioner should validate the assumption and manually annotate several counterfactual instances before conducting the analysis.

\paragraph{Scope}
The second limitation is the scope of our method.
Since \accro{} tests the attentiveness of models, and expecting the prediction on the counterfactual to change, we only include instances where the original predictions are non-neutral for computing attentiveness.
This leaves cases where a model that achieves a good score on \accro{} may be inattentive to partial inputs in those cases.
Therefore, evidence of attentiveness cannot rule out inattentiveness on neutral instances, and similarly evidence of inattentiveness may be even more serious when originally-neutral instances are included.
We leave it to future work to automatically estimate \adj{} for all possible instances.

\section*{Acknowledgments}
We want to thank Alisa Liu, Sofia Serrano, Marius Mosbach, and Shauli Ravfogel for discussions and feedback on this project and this draft.
This work was supported in part by the National Science Foundation (NSF) grants 2007398, 2217154, and NSF IIS-2046873.

\bibliography{custom,anthology}

\clearpage
\newpage

\appendix

\section{Models' Standard Performance}
\label{app:standard-acc}

\begin{table*}[ht]
\centering
\begin{adjustbox}{max width=\textwidth}
\begin{tabular}{@{} lrrr  m{0.01em}  rr  m{0.01em}  rr @{}}
\toprule
& \multicolumn{3}{c}{\bf  NLI} && \multicolumn{2}{c}{ \bf  PD} && \multicolumn{2}{c}{ \bf RC}\\
\cline{2-4}
\cline{6-7}
\cline{9-10}
{\bf Dataset} & \bf RTE & \bf  MNLI & \bf  WANLI && \bf  QQP & \bf  PAWS && \bf  SQuAD &\bf  NewsQA \\

\midrule
BERT-base     &  65.3 &  83.5 &   68.6 &&  91.0 &  90.7 &&   74.1 &    59.4 \\
BERT-large    &  70.0 &  86.3 &   69.9 &&  91.5 &  93.3 &&   80.5 &    61.8 \\
RoBERTa-base  &  71.1 &  86.8 &   72.7 &&  91.6 &  94.4 &&   80.3 &    61.7 \\
RoBERTa-large &  79.8 &  90.1 &   75.4 &&  92.1 &  95.7 &&   84.9 &    65.7 \\
DeBERTa-base  &  76.5 &  90.0 &   75.4 &&  92.4 &  95.6 &&   87.2 &    65.1 \\
DeBERTa-large &  87.7 &  91.3 &   76.7 &&  93.0 &  95.7 &&   89.6 &    66.0 \\
T5-base       &  67.5 &  85.8 &   70.2 &&  91.5 &  93.9 &&   70.8 &    61.0 \\
T5-large      &  87.7 &  89.3 &   76.2 &&  92.1 &  94.9 &&   87.5 &    62.4 \\
T5-1.1-base   &  45.5 &  60.9 &   49.8 &&  87.7 &  55.8 &&   78.9 &    59.7 \\
T5-1.1-large  &  52.7 &  90.5 &   71.3 &&  90.3 &  92.6 &&   88.7 &    61.7 \\
FLAN-T5-base  &  79.8 &  86.1 &   71.2 &&  91.7 &  94.0 &&   83.3 &    60.5 \\
FLAN-T5-large &  89.5 &  90.4 &   76.4 &&  91.0 &  95.1 &&   89.4 &    63.3 \\
\bottomrule
\end{tabular}

\end{adjustbox}
\caption{Standard evaluation performance on respective datasets.}
\label{tab:standard-perf}
\end{table*}
\begin{table*}[t]
\centering
\begin{tabular}{@{} l crrr m{0.01em} rr @{}}
\toprule
& \multicolumn{4}{c}{\bf NLI} 
&& \multicolumn{2}{c}{\bf PD} \\ 
\cline{3-5}
\cline{7-8}
\textbf{Model} & \textbf{\# Tuples}  & \textbf{RTE} & \textbf{MNLI} & \textbf{WANLI} & &\textbf{QQP} & \textbf{PAWS}   \\
\midrule

Majority      & - & 52.7 & 35.4  & 48.3 &&  63.2 &  55.8 \\

\midrule 
GPT3 & 0
& 74.7
& 53.7
& 58.3
&&
50.5
& 71.7
\\

GPT3 & 1
& 86.3
& 74.9
& 55.8
&&
 80.0
& 77.6
\\

GPT3 & 2
& 86.3
& 76.3
& 58.0
&&
 81.0
& 77.3
\\

GPT3 & 3
& 88.1
& 77.4
& 58.1
&&
 81.5
& 78.1
\\

\midrule
\midrule

Flan-T5 Small & 0
& 47.7
& 38.1
& 24.3
&&
 63.2
& 55.7
\\

Flan-T5 Small & 1
& 44.4
& 43.5
& 35.7
&& 
 63.3
& 49.8
\\

Flan-T5 Small & 2
& 44.4
& 35.4
& 31.4
&&
 63.4
& 51.4
\\

Flan-T5 Small & 3
& 50.5
& 34.5
& 31.5
&&
 63.3
& 54.2
\\

\midrule

Flan-T5 Base & 0
& 75.5
& 73.5
& 48.3
&&
 70.8
& 85.8
\\

Flan-T5 Base & 1
& 74.7
& 71.4
& 46.2
&&
 68.6
& 83.4
\\

Flan-T5 Base & 2
& 69.3
& 36.5
& 38.1
&&
 68.4
& 79.2
\\

Flan-T5 Base & 3
& 50.5
& 32.3
& 34.8
&&
 68.2
& 51.5
\\

\midrule

Flan-T5 Large & 0
& 88.1
& 86.8
& 58.7
&&
84.9
& 91.3
\\

Flan-T5 Large & 1
& 88.4
& 86.3
& 59.1
&& 
 84.4
& 90.6
\\

Flan-T5 Large & 2
& 75.8
& 37.5
& 40.7
&&
 84.5
& 86.2
\\

Flan-T5 Large & 3
& 50.5
& 32.1
& 37.8
&&
 84.4
& 52.5
\\

\midrule

Flan-T5 XL & 0
& 89.5
& 89.5
& 61.7
&& 85.9
& 94.4
\\

Flan-T5 XL & 1
& 89.2
& 89.2
& 62.0
&& 85.6
& 94.2
\\

Flan-T5 XL & 2
& 76.5
& 37.8
& 42.0
&&
85.6
& 88.8
\\

Flan-T5 XL & 3
& 50.5
& 32.2
& 35.1
&&
85.5
& 53.2
\\

\midrule

T5 Small & 0
& 50.5
& 80.4
& 52.7
&&
88.4
& N/A
\\

T5 Small & 1
& 47.3
& 42.0
& 35.7
&&
70.3
& N/A
\\

T5 Small & 2
& 47.3
& 38.1
& 34.2
&&
 65.7
& N/A
\\

T5 Small & 3
& 47.3
& 34.5
& 32.0
&&
 63.4
& N/A
\\

\midrule

T5 Base & 0
& 67.5
& 85.6
& 59.0
&&
90.6
& N/A
\\

T5 Base & 1
& 49.1
& 40.1
& 33.2
&&
69.0
& N/A
\\

T5 Base & 2
& 51.3
& 37.9
& 37.1
&&
66.4
& N/A
\\

T5 Base & 3
& 49.5
& 35.1
& 36.0
&&
 64.4
& N/A
\\

\midrule

T5 Large & 0
& 85.9
& 89.7
& 61.7
&&
91.3
& N/A
\\

T5 Large & 1
& 82.7
& 42.7
& 36.6
&&
74.9
& N/A
\\

T5 Large & 2
& 75.8
& 40.2
& 33.3
&&
 70.9
& N/A
\\

T5 Large & 3
& 57.8
& 35.7
& 25.7
&&
 67.4
& N/A
\\

\midrule

T5 3B & 0
& 87.7
& 91.1
& 63.3
&& 90.6
& N/A
\\

T5 3B & 1
& 52.0
& 40.3
& 31.3
&&   68.6
& N/A
\\

T5 3B & 2
& 47.3
& 36.3
& 21.2
&&  64.7
& N/A
\\

T5 3B & 3
& 43.0
& 33.7
& 16.4
&& 63.3
& N/A
\\

\bottomrule
\end{tabular}
\caption{In-context learning dev. set accuracy. T5 models fail to output valid labels on PAWS.}
\label{tab:icl_acc}
\end{table*}
\begin{table}[h]
\centering
\begin{adjustbox}{max width=\columnwidth}
\begin{tabular}{lrr}
\toprule

\bf Dataset & \bf VQA v2 & \bf NLVR2 \\
\midrule

 BLIP & 77.5 & 81.1 \\
 ViLT & 70.3 & 74.6 \\

\bottomrule
\end{tabular}
\end{adjustbox}
\caption{Standard performance on the VLR tasks benchmarked on the finetuned model with full-inputs}
\label{tab:vr-vqa-nlvr}
\end{table}
\begin{table*}[ht]
\centering
\begin{tabular}{@{} l c rrr m{0.001em} rr   @{}} %

\toprule
&& \multicolumn{3}{c}{\textbf{NLI}} && \multicolumn{2}{c}{\textbf{PD}}  \\ 
\cline{3-5}
\cline{7-8}
\bf Model & \bf \#Tuples & \bf RTE & \bf MNLI & \bf WANLI && \bf QQP &  \bf PAWS\\
\midrule

GPT3 & 0
& 96.3$_{\pm 0.9 }$
& 94.6$_{\pm 0.1 }$
& 94.3$_{\pm 0.3 }$
&
& 97.2$_{\pm 0.1 }$
& 100.0$_{\pm 0.0 }$
\\
GPT3 & 1
& 100.0$_{\pm 0.0 }$
& 31.3$_{\pm 0.3 }$
& 36.7$_{\pm 0.4 }$
&
& 100.0$_{\pm 0.0 }$
& 100.0$_{\pm 0.0 }$
\\
GPT3 & 2
& 100.0$_{\pm 0.0 }$
& 26.0$_{\pm 0.7 }$
& 41.0$_{\pm 0.8 }$
&
& 100.0$_{\pm 0.0 }$
& 100.0$_{\pm 0.0 }$
\\
GPT3 & 3
& 100.0$_{\pm 0.0 }$
& 25.6$_{\pm 0.5 }$
& 41.8$_{\pm 0.9 }$
&
& 100.0$_{\pm 0.0 }$
& 100.0$_{\pm 0.0 }$
\\

\midrule\midrule
Flan-T5-base & 0
& 100.0$_{\pm 0.0 }$
& 55.9$_{\pm 0.2 }$
& 57.9$_{\pm 0.3 }$
&
& 99.7$_{\pm 0.0 }$
& 99.9$_{\pm 0.0 }$
\\
Flan-T5-base & 1
& 98.6$_{\pm 1.1 }$
& 23.5$_{\pm 0.7 }$
& 23.9$_{\pm 0.6 }$
&
& 99.7$_{\pm 0.0 }$
& 99.9$_{\pm 0.0 }$
\\\midrule
Flan-T5-large & 0
& 100.0$_{\pm 0.0 }$
& 94.1$_{\pm 0.1 }$
& 94.3$_{\pm 0.3 }$
&
& 100.0$_{\pm 0.0 }$
& 100.0$_{\pm 0.0 }$
\\
Flan-T5-large & 1
& 99.9$_{\pm 0.2 }$
& 95.8$_{\pm 0.6 }$
& 96.5$_{\pm 0.2 }$
&
& 100.0$_{\pm 0.0 }$
& 100.0$_{\pm 0.0 }$
\\\midrule
Flan-T5-XL & 0
& 100.0$_{\pm 0.0 }$
& 99.0$_{\pm 0.1 }$
& 99.2$_{\pm 0.1 }$
&
& 100.0$_{\pm 0.0 }$
& 100.0$_{\pm 0.0 }$
\\
Flan-T5-XL & 1
& 100.0$_{\pm 0.0 }$
& 98.5$_{\pm 0.5 }$
& 99.3$_{\pm 0.1 }$
&
& 100.0$_{\pm 0.0 }$
& 100.0$_{\pm 0.0 }$
\\

\midrule

T5-base & 0
& 100.0$_{\pm 0.0 }$
& 72.6$_{\pm 0.2 }$
& 87.6$_{\pm 0.3 }$
&
& 99.9$_{\pm 0.0 }$
& -
\\
T5-base & 1
& 99.8$_{\pm 0.3 }$
& 53.7$_{\pm 2.6 }$
& 57.6$_{\pm 7.8 }$
&
& 97.9$_{\pm 0.4 }$
& -
\\\midrule
T5-large & 0
& 100.0$_{\pm 0.0 }$
& 65.9$_{\pm 0.2 }$
& 77.6$_{\pm 0.3 }$
&
& 100.0$_{\pm 0.0 }$
& -
\\
T5-large & 1
& 99.0$_{\pm 0.8 }$
& 43.0$_{\pm 4.0 }$
& 37.2$_{\pm 8.9 }$
&
& 98.8$_{\pm 0.3 }$
& -
\\\midrule
T5-3B & 0
& 100.0$_{\pm 0.0 }$
& 61.7$_{\pm 0.2 }$
& 72.7$_{\pm 0.4 }$
&
& 100.0$_{\pm 0.0 }$
& -
\\
T5-3B & 1
& 100.0$_{\pm 0.0 }$
& 27.3$_{\pm 3.3 }$
& 28.6$_{\pm 1.9 }$
&
& 98.8$_{\pm 0.4 }$
& -
\\

\bottomrule
\end{tabular}
\caption{Accuracy of in-context learning models on the standard dev. split without perturbation, averaged across five different random seeds.
T5 models fail to output valid labels on PAWS.
The small-sized T5 model, in the three-shot setting, always makes neutral predictions on the original RTE dev. set, making its counterfactual accuracy unavailable.
}
\label{tab:icl_counterfactual-acc}
\end{table*}

We report the majority-class accuracy and the full results of the supervised models (BERT, RoBERTa, DeBERTa, both versions of T5 and Flan-T5) in \autoref{tab:standard-perf} for NLI and PD. The results of the models used in few-shot in-context learning experiments (both versions of T5, Flan-T5 and GPT-3) are reported in \autoref{tab:icl_acc}.

We highlight a few interesting trends. First, DeBERTa-large and Flan-T5-large obtain similar performance across the tasks and datasets.
Second, while T5 v1 typically performs better across tasks and models' sizes than v1.1 - which is expected as the training data of many of these dataset were part of training data of the later - it is especially expected noticable on RTE, a very small dataset (2,490 instances for training) - and in the smaller models. For instance, T5-small performs 67.87\% on RTE, while the corresponding v1.1 model only achieves 47.29\%, which is worse than random.

The ICL results can be found in Appendix \ref{app:standard-acc} in Tables \ref{tab:standard-perf}, \ref{tab:vr-vqa-nlvr}, \ref{tab:icl_acc}; they illustrate two opposing trends. GPT-3 benefits from more demonstrations, where the biggest benefits happen with three-shots, and additional demonstrations' benefit is more subtle. For instance, GPT-3 achieves 53.7\% accuracy on MNLI in zero-shot, then 74.9\% with 3-shots and 77.4\% with 9-shots.
On the other hand, the performance of both the T5 and Flan-T5 models, across all sizes (from small to XL), performs worse the more demonstrations are shown.\footnote{Except for Flan-T5 Small that gets worse than random.} For instance, on MNLI Flan-T5-XL achieves 89.5\% accuracy, slightly decreases to 89.2 with 3-shot, and then completely fails with 6-, and 9-shots (37.8\% and 32.3\%, respectively).
While \citet{min2022rethinking} found that demonstrations contribute by providing examples of the label space, input distribution, and format, and are not as attentive to the content itself, we find that demonstrations harm models in performing the task (except for GPT-3). 

In the PD datasets, we notice that the performance of all models are rather high, and the gap between all models is small (5.37 in QQP).

\section{Partial Input Baseline}
\label{app:partial-input-baseline}
\begin{table}[h]
\centering
\begin{adjustbox}{max width=\columnwidth}
\begin{tabular}{llrr}
\toprule

\bf  Setup & \bf Model/Data & \bf VQA v2 & \bf NLVR2 \\
\midrule

\multirow{2}{*}{Image-only}
 & BLIP & 24.2 & 50.9 \\
 & ViLT & 25.4 & 49.3 \\

\midrule

\multirow{2}{*}{Text-only}
 & BLIP & 40.0 & 42.2 \\
 & ViLT & 50.7 & 49.3 \\

\bottomrule
\end{tabular}
\end{adjustbox}
\caption{Standard performance on the Visual Reasoning tasks benchmarked on partial inputs.}
\label{tab:vr-partial}
\end{table}
\begin{table*}[ht]
\centering
\begin{adjustbox}{max width=\textwidth}
\begin{tabular}{@{} l  rrr  m{0.01em}  rr  m{0.01em}  rrr  m{0.01em}  rrr @{}}
\toprule
& \multicolumn{3}{c}{\bf NLI} && \multicolumn{2}{c}{\bf PD} && \multicolumn{3}{c}{\bf RC - Question Only} && \multicolumn{3}{c}{\bf RC - Paragraph Only} \\
\cline{2-4}
\cline{6-7}
\cline{9-11}
\cline{13-15}
\bf Model/Data & \bf   RTE & \bf  MNLI & \bf  WANLI && \bf   QQP & \bf  PAWS && \bf  SQuAD & \bf DuoRC & \bf  NewsQA &&\bf   SQuAD & \bf DuoRC & \bf   NewsQA \\
\midrule
Majority      & 52.7  &  35.4 &   48.3 &&  63.2 &  55.8 && 33.4 & & 28.8 && 33.4 & & 28.8 \\
\midrule
BERT-base     &  56.0 &  56.6 &   50.8 &&  77.9 &  56.6 &&     95.6 &     67.2 &      90.7 &&     50.1 &      6.1 &      27.5 \\
BERT-large    &  57.4 &  57.1 &   50.7 &&  77.5 &  56.2 &&     95.5 &     69.7 &      91.8 &&     50.1 &      6.3 &      27.6 \\
RoBERTa-base  &  52.0 &  59.3 &   53.3 &&  77.9 &  58.1 &&     95.1 &     70.2 &      96.3 &&     50.1 &      6.6 &      27.5 \\
RoBERTa-large &  56.3 &  59.7 &   51.7 &&  77.7 &  55.8 &&     96.5 &     72.2 &      96.7 &&     50.1 &      6.8 &      27.5 \\
DeBERTa-base  &  61.4 &  60.0 &   52.7 &&  76.5 &  58.0 &&     98.0 &     70.9 &      96.6 &&     50.1 &      7.0 &      27.5 \\
DeBERTa-large &  61.4 &  61.1 &   52.4 &&  78.3 &  58.4 &&     98.5 &     73.2 &      96.8 &&     50.1 &      7.0 &      27.6 \\
T5-base       &  56.0 &  57.9 &   50.1 &&  78.8 &  58.6 &&     93.5 &     54.0 &      61.2 &&     50.2 &      5.9 &      27.5 \\
T5-large      &  63.2 &  61.2 &   54.7 &&  77.8 &  56.9 &&     97.1 &     58.4 &      64.3 &&     50.1 &      6.9 &      27.5 \\
T5-1.1-base   &  56.3 &  56.4 &   49.0 &&  77.9 &  56.1 &&     95.1 &     54.8 &      62.3 &&     50.1 &      5.9 &      27.5 \\
T5-1.1-large  &  53.1 &  61.7 &   49.3 &&  74.3 &  53.9 &&     97.7 &     58.5 &      64.1 &&     50.1 &      6.8 &      27.5 \\
FLAN-T5-base  &  53.8 &  57.6 &   50.9 &&  78.6 &  58.0 &&     95.2 &     56.4 &      62.0 &&     50.1 &      6.5 &      27.5 \\
FLAN-T5-large &  59.2 &  61.5 &   55.7 &&  75.5 &  56.3 &&     97.5 &     58.5 &      64.4 &&     50.1 &      7.0 &      27.5 \\
\bottomrule
\end{tabular}

\end{adjustbox}
\caption{Partial-input baselines on NLI, PD, and RC.}
\label{tab:partial-inputs}
\end{table*}

We report the full results of the partial input baselines in Tables \ref{tab:vr-partial}, \ref{tab:partial-inputs}.

In NLI, we train models on the \textit{hypotheses} (or \textit{sentence1} in RTE), \textit{sentence1} in PD, either on the \textit{question} or \textit{paragraph} in RC,\footnote{Following \citet{kaushik-lipton-2018-much} it is implemented by randomly sampling a different passage from the corpus and adding the answer candidate to this passage in a random location. %
} and either \textit{image} or the \textit{text} in VLR.

On QQP, T5-base trained on the corresponding dataset achieves 78.81\% (compared to the 63.18\% majority accuracy).

Similarly, models finetuned on VLR datasets perform much higher than the majority class (24.2\%) 
However, while image-only and text-only models perform similarly when finetuned for NLVR2, we observe that the text-only performance is much higher for VQA v2 dataset (39.9\% for BLIP and 42.1\% for ViLT) than the image-only ($\sim$ 24\%). The image-only model performs close to the majority class baseline. This demonstrates unequal correlations in different modalities.

\begin{table*}[ht!]
\centering
\begin{tabularx}{\textwidth}{@{} l X  @{}}
    \toprule
    \textbf{Task} & \textbf{Instruction}  \\
    \midrule
    
    \multirow{4}{*}{RTE} & 
    You're given a pair of sentences: a Text and a Hypothesis. 
    Your job is to determine the relation between them based on your inference from the statement and your commonsense knowledge. 
    Answer `Entailment' if the Hypothesis can be inferred from the Text;
    Answer `Not entailment' if the Hypothesis disagrees with the Text.\\

    \midrule
    
     \multirow{6}{*}{MNLI, WANLI} & You're given a pair of sentences: a Premise and a Hypothesis. Your job is to determine the relation between them based on your inference from the statement and your commonsense knowledge.
    Answer `Entailment' if the Hypothesis can be inferred from the Premise;
    Answer 'Contradiction' if the Hypothesis disagrees with the Premise
    Answer 'Neutral' if the Hypothesis can neither be inferred from the Premise nor disagrees with the Premise.\\

    \midrule
    
     \multirow{3}{*}{QQP} & You're given a pair of questions. Your job is to determine whether they are semantically equivalent.
    Answer `Paraphrase if they bear the same meaning;
    Answer `Not paraphrase' if they have different meanings.\\
    
    \midrule
    
    \multirow{3}{*}{PAWS} & You're given a pair of sentences. Your job is to determine whether they are semantically equivalent.
           Answer `Paraphrase' if they bear the same meaning;
           Answer `Not paraphrase' if they have different meanings. \\
    \bottomrule
\end{tabularx}
\caption{Textual instructions for the instruction-fintuned models in the in-context learning experiments.
}
\label{tab:icl_prompts}
\end{table*}

\section{Full \accro{} Results}

Here we provide the full results for all datasets we consider.
The results are summarized in Table \ref{tab:counterfactuals-dev}.

\begin{table*}[ht]
\begin{adjustbox}{max width=\textwidth}
\centering
\begin{tabular}{@{} lrrr  m{0.01em}  rr  m{0.01em}  rrr @{}}
\toprule
 & \multicolumn{3}{c}{\bf  NLI} && \multicolumn{2}{c}{ \bf  PD} && \multicolumn{3}{c}{ \bf RC}\\
\cline{2-4}
\cline{6-7}
\cline{9-11}
{\bf Dataset} & \bf RTE & \bf  MNLI & \bf  WANLI && \bf  QQP & \bf  PAWS && \bf  SQuAD & \bf DuoRC & \bf  NewsQA \\
\midrule

BERT-base & 99.8$_{ \pm 0.3}$ & 74.6$_{ \pm 0.3}$ & 97.8$_{ \pm 0.2}$ &  & 99.9$_{ \pm 0.0}$ & 99.3$_{ \pm 0.2}$ &  & 98.0$_{ \pm 0.0}$ & 97.4$_{ \pm 0.1}$ & 99.0$_{ \pm 0.1}$ \\
BERT-large & 100.0$_{ \pm 0.0}$ & 64.3$_{ \pm 0.1}$ & 99.8$_{ \pm 0.1}$ &  & 99.9$_{ \pm 0.0}$ & 85.9$_{ \pm 0.3}$ &  & 98.0$_{ \pm 0.2}$ & 97.6$_{ \pm 0.1}$ & 99.1$_{ \pm 0.1}$ \\
RoBERTa-base & 99.8$_{ \pm 0.3}$ & 73.4$_{ \pm 0.5}$ & 99.5$_{ \pm 0.1}$ &  & 99.9$_{ \pm 0.0}$ & 80.5$_{ \pm 0.6}$ &  & 98.3$_{ \pm 0.1}$ & 97.7$_{ \pm 0.1}$ & 99.1$_{ \pm 0.2}$ \\
RoBERTa-large & 100.0$_{ \pm 0.0}$ & 69.8$_{ \pm 0.2}$ & 99.6$_{ \pm 0.1}$ &  & 99.9$_{ \pm 0.0}$ & 78.7$_{ \pm 0.3}$ &  & 98.2$_{ \pm 0.2}$ & 98.1$_{ \pm 0.1}$ & 99.1$_{ \pm 0.1}$ \\
DeBERTa-base & 97.0$_{ \pm 1.6}$ & 78.8$_{ \pm 0.4}$ & 99.5$_{ \pm 0.1}$ &  & 99.9$_{ \pm 0.0}$ & 99.9$_{ \pm 0.1}$ &  & 97.6$_{ \pm 0.1}$ & 97.6$_{ \pm 0.3}$ & 99.0$_{ \pm 0.1}$ \\
DeBERTa-large & 100.0$_{ \pm 0.0}$ & 68.6$_{ \pm 0.3}$ & 99.4$_{ \pm 0.1}$ &  & 99.9$_{ \pm 0.0}$ & 99.1$_{ \pm 0.1}$ &  & 98.0$_{ \pm 0.2}$ & 97.9$_{ \pm 0.1}$ & 98.9$_{ \pm 0.1}$ \\
T5-base & 99.4$_{ \pm 0.7}$ & 81.3$_{ \pm 0.1}$ & 99.0$_{ \pm 0.2}$ &  & 99.9$_{ \pm 0.0}$ & 99.0$_{ \pm 0.1}$ &  & 99.1$_{ \pm 0.1}$ & 96.5$_{ \pm 0.1}$ & 99.5$_{ \pm 0.0}$ \\
T5-large & 100.0$_{ \pm 0.0}$ & 71.5$_{ \pm 0.4}$ & 99.8$_{ \pm 0.1}$ &  & 99.9$_{ \pm 0.0}$ & 100.0$_{ \pm 0.0}$ &  & 99.5$_{ \pm 0.1}$ & 97.8$_{ \pm 0.2}$ & 99.7$_{ \pm 0.0}$ \\
T5-1.1-base & - & 46.9$_{ \pm 0.6}$ & - &  & 97.1$_{ \pm 0.1}$ & - &  & 99.4$_{ \pm 0.0}$ & 96.9$_{ \pm 0.1}$ & 99.0$_{ \pm 0.1}$ \\
T5-1.1-large & - & 79.4$_{ \pm 0.3}$ & 99.8$_{ \pm 0.1}$ &  & 99.9$_{ \pm 0.0}$ & - &  & 99.4$_{ \pm 0.1}$ & 97.6$_{ \pm 0.0}$ & 99.6$_{ \pm 0.1}$ \\
FLAN-T5-base & 99.9$_{ \pm 0.3}$ & 81.8$_{ \pm 0.3}$ & 99.5$_{ \pm 0.1}$ &  & 99.9$_{ \pm 0.0}$ & 98.3$_{ \pm 0.2}$ &  & 99.4$_{ \pm 0.1}$ & 97.4$_{ \pm 0.0}$ & 99.4$_{ \pm 0.2}$ \\
FLAN-T5-large & 100.0$_{ \pm 0.0}$ & 77.1$_{ \pm 0.2}$ & 99.8$_{ \pm 0.1}$ &  & 99.9$_{ \pm 0.1}$ & 100.0$_{ \pm 0.1}$ &  & 99.5$_{ \pm 0.1}$ & 98.2$_{ \pm 0.1}$ & 99.7$_{ \pm 0.0}$ \\

\bottomrule
\end{tabular}

\end{adjustbox}
\caption{\accro{} results on natural language inference (NLI), paraphrase detection (PD) and reading comprehension (RC). Results are calculated by computing the mean attentiveness of five counterfactuals per instance, and their standard deviation. Higher numbers indicate better model attentiveness. 
We mark results for models whose standard performance is not significantly better than random with `-'.
}
\label{tab:counterfactuals-dev}
\end{table*}

\section{Comparable Counterfactuals}
\label{app:comparable}
\begin{table}[h]
\centering
\begin{adjustbox}{max width=\columnwidth}
\begin{tabular}{@{} lrrr @{}}
\toprule
\bf Dataset &  \bf RTE &  \bf MNLI &  \bf WANLI \\
\midrule

BERT-base     &   93.3 &  70.8 &   98.4 \\
BERT-large    &  100.0 &  60.2 &  100.0 \\
RoBERTa-base  &  100.0 &  68.8 &   99.9 \\
RoBERTa-large &  100.0 &  64.7 &   99.4 \\
DEBERTA-base  &  100.0 &  74.7 &   99.4 \\
DeBERTa-large &  100.0 &  64.5 &   99.3 \\
T5-base       &   93.3 &  77.5 &   98.9 \\
T5-large      &  100.0 &  67.2 &  100.0 \\
T5-1.1-base   &  -     &  43.3 &   -    \\
T5-1.1-large  &  -     &  74.6 &  100.0 \\
FLAN-T5-base  &  100.0 &  78.3 &   99.4 \\
FLAN-T5-large &  100.0 &  72.8 &  100.0 \\

\bottomrule
\end{tabular}
\end{adjustbox}
\caption{Comparable partial-inputs counterfactuals on NLI datasets.}
\label{tab:comparable-nli-dev}
\end{table}

In Section \ref{sec:counterfactuals} we report the results of each model based on its own original predictions. As a result, the models evaluated on the same dataset are not entirely comparable since they are evaluated on different subsets. As such, in this section we perform the evaluation on the same subsets across models, to perform a fair comparison. We report the results in Table \ref{tab:comparable-nli-dev}. The average difference between each model between the main results and the comparable ones on the same subset is minimal. 
This results suggest that even if the models are evaluated on different subsets, it is enough to gain insights about the inspected models.

\section{Comparison with \texorpdfstring{\citet{srikanth-rudinger-2022-partial}}{Srikanth and Rudinger (2022)}}\label{sec:comparison}
We provide additional comparison with \citet{srikanth-rudinger-2022-partial} in Table \ref{tab:srikanth-comparison}.

\begin{table}[h]
\begin{adjustbox}{max width=\columnwidth}
\begin{tabular}{@{} lll @{}}
\toprule
\bf Property & \bf \citet{srikanth-rudinger-2022-partial} & \bf Our Work \\
\midrule
Tasks & NLI & NLI, PD, RC, V\&L-R \\
NLI-Datasets & SNLI, $\delta$-NLI & RTE, MNLI, WANLI \\
Models & RoBERTa-base & BERT*, RoBERTa*, DeBERTa*, \\
& & T5* T5-v1.1*, Flan-T5*, GPT-3 \\
Counterfactuals & Manual & Automatic \\
Examples subset & Heuristic prone & Non-neutral prediction \\
Epochs & 2 & 10 \\
New labels & All & Neutral \\
Conclusions & No reliance & Data \& Model dependent \\

\bottomrule
\end{tabular}

\end{adjustbox}
\caption{Summarizing our setup compared to \citet{srikanth-rudinger-2022-partial}. * signifies we test several model sizes from the same family.}
\label{tab:srikanth-comparison}
\end{table}

\section{Responsible Research Checklist}

\paragraph{License}
All datasets we use use the cc-by-3.0 license, besides PAWS, which has a different license: \url{https://github.com/google-research-datasets/paws/blob/master/LICENSE}.

\paragraph{Compute}
To run experiments we used nvidia A-100 gpus, for an estimated time of 100 hours.

\end{document}